\documentclass[letterpaper, 10 pt, conference]{ieeeconf}  
\usepackage{cite}
\usepackage{amsmath,amssymb,amsfonts}
\usepackage{algorithmic}
\usepackage{graphicx,color}
\usepackage{textcomp}
\usepackage{siunitx}
\usepackage{booktabs}
\usepackage{tabularx} 
\usepackage{subcaption}
\usepackage{wrapfig}
\usepackage{hyperref}
\usepackage{amsmath}
\usepackage{hyperref}
\usepackage{array}
\usepackage{xcolor}    
\newcommand{\acro}[1]{\textcolor{black}{\textbf{#1}}} 

\IEEEoverridecommandlockouts                              

\IEEEoverridecommandlockouts                              

\title{%
\acro{M}\acro{V}\acro{P}-\acro{Tac}: A \acro{M}iniaturized Dual-Modal \acro{V}ision and \acro{P}hotoelastic \acro{Tac}tile Sensor for Robot-Assisted Minimally Invasive Surgery%
}

\author{
Md Rakibul Islam Prince$^{1}$,
Jaeeun Kim$^{2}$,
Yuhao Zhou$^{2}$,
Mason Vrshek$^{3}$,
Shivani Reddy Sama$^{1}$,
Adyaa Khera$^{4}$, \\
Sheeraz Athar$^{2}$,
Zijie Xu$^{5}$,
Jiabin Liu$^{6}$,
Shaoting Lin$^{6}$,
Wei Li$^{5}$,
and Yu She$^{1,2,3}$%
\thanks{*This work was supported in part by the Department of the Air Force under Contract FA857126C0002; in part by the National Science Foundation under Awards 2423068 and 2520136; and in part by the U.S. Department of Agriculture under Awards 2023-67021-39072 and 2024-67021-42878; and in part by Showalter Trust.}%
\thanks{$^{1}$Elmore Family School of Electrical and Computer Engineering, Purdue University, West Lafayette, IN 47907, USA. \tt\small{\{prince26, sama, shey\}@purdue.edu}}%
\thanks{$^{2}$Edwardson School of Industrial Engineering, Purdue University, West Lafayette, IN 47907, USA. \tt\small{\{kim2592, zhou1437, sathar, shey\}@purdue.edu}}%
\thanks{$^{3}$School of Mechanical Engineering, Purdue University, West Lafayette, IN 47907, USA. \tt\small{\{mvrshek, shey\}@purdue.edu}}%
\thanks{$^{4}$Robotics Engineering Technology, Purdue University, West Lafayette, IN 47907, USA. \tt\small khera2@purdue.edu}%
\thanks{$^{5}$Department of Civil Engineering, Stony Brook University, Stony Brook, NY, USA. \tt\small{\{zijie.xu, wei.li.6\}@stonybrook.edu}}%
\thanks{$^{6}$Department of Mechanical Engineering, Michigan State University, East Lansing, MI, USA. \tt\small{\{liujiab1, linshaot\}@msu.edu}}%
}

\begin{document}

\maketitle
\thispagestyle{empty}
\pagestyle{empty}

\begin{abstract}
Robot-assisted minimally invasive surgery (RMIS) offers major benefits over open and conventional laparoscopic procedures, yet it still lacks tactile feedback for palpation while operating under strict requirements to preserve reliable vision for navigation and safety. In practice, visual feedback is indispensable, and tactile solutions that cannot coexist with vision are difficult to translate into RMIS tools. To address both needs, we introduce MVP-Tac, a compact, vision-based tactile sensor that provides co-located vision and tactile sensing. MVP-Tac uses reflective photoelastic imaging: a thin photoelastic elastomer produces stress-dependent interferograms under contact that are captured by an embedded camera through a miniaturized reflective polariscope. A semi-transparent membrane and controllable illumination enable switching between visual mode and tactile mode, enabling tactile perception without sacrificing vision. We validate MVP-Tac through force calibration in the 0 to 2~N range and demonstrate its potential for tumor palpation via video-based hardness classification on tissue phantoms, achieving 
97\% accuracy for exposed-tumor classification and 92\% accuracy for subdermal-tumor classification. Finally, we conduct a simulated colonoscopy to validate both visual and tactile modalities in a constrained lumen, including vision-guided 3D photomapping of the luminal wall and in situ hardness classification of localized nodules. Overall, MVP-Tac provides a practical path toward restoring clinically useful palpation in RMIS while maintaining essential visual feedback. The design, fabrication, and firmware of MVP-Tac are open-sourced at
\url{https://mvp-tac.github.io/}.
\end{abstract}

\begin{figure}[!t]
    \centering
    \includegraphics[width=\columnwidth]{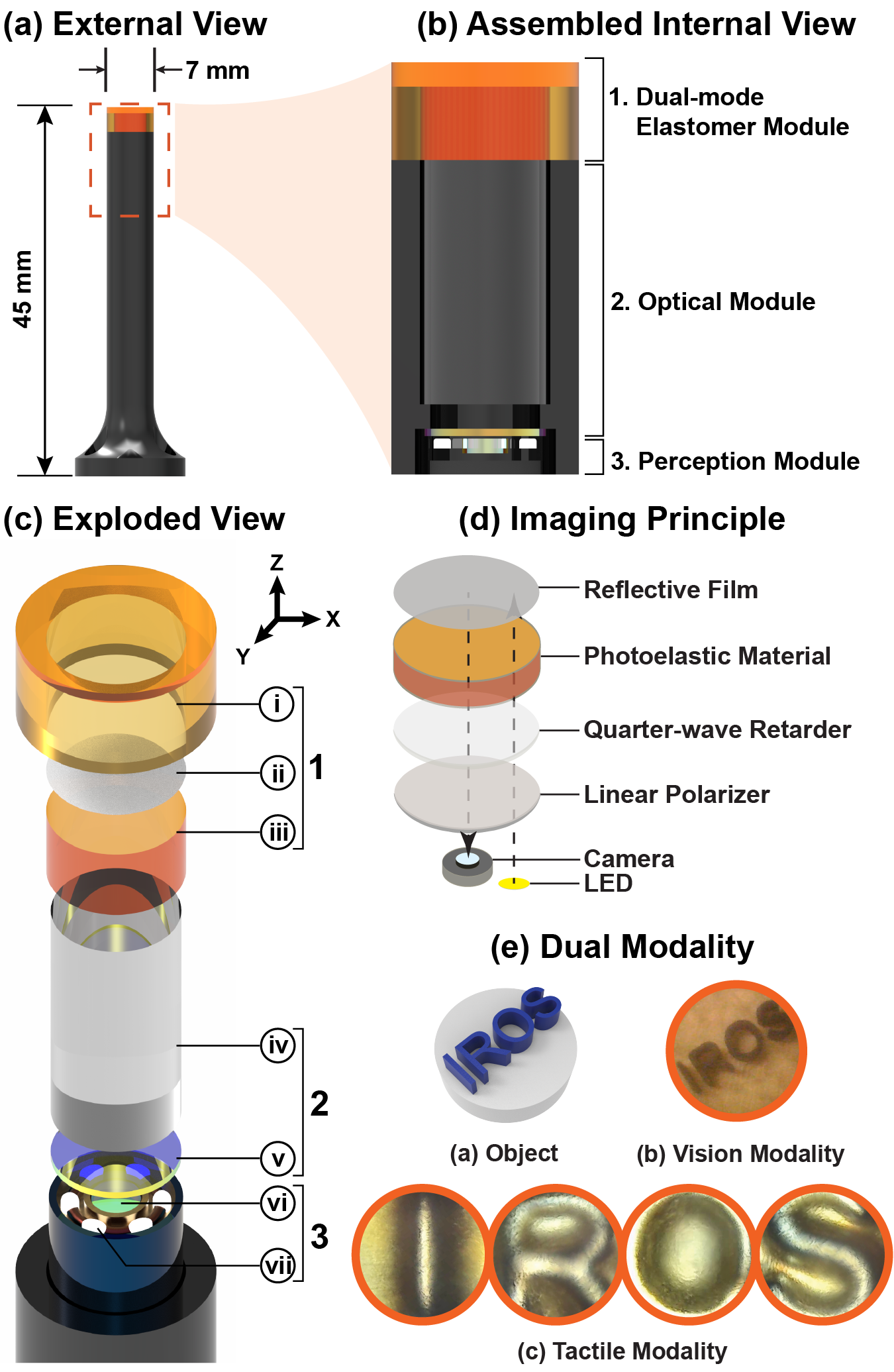}
    \caption{\textbf{Overview of the MVP-Tac sensor.}
(a) External view of the sensor showing its overall form factor and outer dimensions.
(b) Assembled internal view highlighting the arrangement of the main sensing modules.
(c) Exploded view illustrating the key structural and sensing components of the system: (i) Protective elastomer layer, (ii) Semi-transparent reflective coating, (iii) Photoelastic elastomer layer, (iv) Acrylic Spacer, (v) Circular polarizer film, (vi) Camera and (vii) LEDs.
(d) Imaging principle of the sensor based on photoelasticity.
(e) Dual-modality operation of the sensor} 
    \label{fig:sensor}
\end{figure}

\section{INTRODUCTION}

Early and accurate localization of tumors during minimally invasive and endoscopic procedures remains a significant clinical challenge to this day. In open surgery, surgeons rely heavily on manual palpation to detect stiffness variations associated with malignant masses, particularly for small or subsurface tumors that may not be visually distinguishable. However, in robot-assisted minimally invasive surgery (RMIS) and flexible endoscopy, direct tactile feedback is largely absent. As a result, surgeons must depend primarily on visual cues from endoscopic imaging modalities, which are often insufficient for detecting deeply embedded or mechanically distinct lesions.

Current intraoperative tumor detection methods rely predominantly on visual inspection, preoperative imaging correlation, intraoperative ultrasound, or optical enhancement techniques. Existing modalities such as endoscopic ultrasound (EUS) require additional instrumentation, specialized training, and interpretation and may not offer the spatial resolution or intuitive mechanical feedback of direct palpation. Additionally, visual-only endoscopy cannot reliably identify small, stiff inclusions beneath the compliant tissue when the surface morphology appears normal. As a result, the absence of haptic feedback in RMIS and endoscopy can lead to missed lesions, incomplete resections, or overly aggressive tissue exploration.

To address these challenges, we introduce MVP-Tac, a novel, vision-based tactile sensor leveraging photoelasticity for advanced clinical applications. The sensor's inferred force-feedback is accompanied by rich visual data, and its compact integration is suitable for endoscopic deployment. MVP-Tac bridges the gap between high-resolution vision-based tactile sensing and clinically viable tumor palpation in RMIS, representing a step toward restoring intuitive mechanical perception in minimally invasive oncology procedures.

The key contributions of this work are as follows.
\begin{itemize}
    \item We present the miniature design, setup, and development of MVP-Tac that provides its rich visual data along with force-feedback for practical implementation in the surgical industry.
    \item We detail our approach that estimates force by utilizing photoelastic material and a polariscope setup, eliminating the need for a structured lighting, making MVP-Tac a compact and clinically practical solution.
    \item We validate MVP-Tac on exposed and subdermal tumor phantoms for hardness classification, and further demonstrate simulated colonoscopy with vision-guided localization, tactile palpation, 3D luminal photomapping, and in situ hardness classification.
\end{itemize}



\section{RELATED WORK}
Vision-based tactile sensors have been developed to encode depth information through controlled illumination gradients or reference frameworks for depth calibration~\cite{10.1145/2010324.1964941, yuan2017gelsight}. The two primary families dominate this area: the GelSight family~\cite{yuan2017gelsight,wang2021gelsight,taylor2022gelslim} and the Marker family~\cite{xue20233, du2021high, yu2019novel}.

GelSight sensors~\cite{yuan2017gelsight} and their variants~\cite{dong2017improved, taylor2022gelslim, athar2025vibtac} use photometric stereo~\cite{johnson2009retrographic} to reconstruct high-resolution, 3D geometry of contact objects. These sensors enable texture recognition~\cite{li2013sensing, luo2018vitac}, dexterous manipulation~\cite{tian2019manipulation, wang2020swingbot,she2021cable}, and shape reconstruction~\cite{bauza2019tactile, hu2025vibrissae, suresh2022shapemap}. However, photometric stereo requires carefully controlled internal illumination~\cite{tippur2023gelsight360}, complex fabrication due to precise LED placement~\cite{romero2020soft} and reflective pigment layers~\cite{yuan2017gelsight}. This internal lighting structure also limits scalability for compact platforms such as RMIS. In contrast, Marker-family sensors estimate depth through binocular vision and optical flow by aligning markers observed by two cameras~\cite{xue20233, yu2019novel}. While dense marker distributions improve gradient representation, they can interfere with feature extraction and reliable marker tracking~\cite{xue20233}.

Several approaches also fall outside of these two families. Darkness mapping sensors~\cite{lin2023dtact, lin20239dtact} infer contact depth by correlating brightness variations with deformation using semi-transparent and opaque layers under uniform illumination, eliminating the need for structured lighting and marker tracking, but the opaque layer prevents simultaneous camera vision, limiting its use in endoscopic settings. Recent work on miniaturized sensors for RMIS~\cite{li2024minitac, prince2025tacscope} introduces compact tactile devices; however, they rely on opaque contact layers—mechanoresponsive photonic materials~\cite{li2024minitac} or dense particle coatings~\cite{prince2025tacscope}—which similarly obstruct camera vision. Photoelasticity offers an alternative by optically encoding internal stress fields through stress-induced birefringence~\cite{aben2012photoelasticity, ramesh2002digital}. Nevertheless, most photoelastic methods require bulky polariscope setups~\cite{daniels2017photoelastic, puckett2013equilibrating, liu2025fatigue} and often report only global intensity measurements (e.g., photodiode-based) rather than spatially resolved contact information~\cite{mitsuzuka2022application}.

To address these limitations, we propose MVP-Tac, a compact vision-based tactile sensor designed for RMIS to preserve endoscopic \emph{vision}. Most tactile sensing approaches do not support co-located vision, and although photometric-stereo sensors can enable see-through operation with semi-transparent coatings, structured illumination is difficult to implement reliably at miniaturized scales. MVP-Tac instead employs reflective photoelastic imaging, which requires no structured lighting and naturally supports reflective coatings. With a semi-transparent, half-silvered membrane and controllable illumination, MVP-Tac enables dual-mode operation for integrated vision and tactile sensing in a compact form-factor. \textbf{To the best of our knowledge, MVP-Tac is the first miniaturized endoscope-compatible tactile sensor to integrate co-located vision and tactile sensing within a single compact package at this scale.} We further develop a calibration and learning pipeline that combines a gradient-based photoelastic metric ($G^2$) with image features for robust force estimation and video-based hardness classification. The effectiveness of MVP-Tac is validated through controlled experiments on tissue-mimicking phantoms for both exposed and subdermal tumor palpation scenarios.

\section{THE MVP-Tac SENSOR}

\subsection{Photoelastic Imaging Principle}
Photoelasticity is an optical phenomenon in which a transparent, compliant material becomes \emph{stress-dependent birefringent} under mechanical loading. As polarized light passes through the stressed material, the induced birefringence introduces a relative phase delay (retardation) that can be imaged and used as a tactile signal.

In MVP-Tac, the stress-induced birefringence is summarized by an optical retardation map along the light propagation direction (the $z$-axis):
\begin{equation}
R_t(x,y)=C_{so}\int_{0}^{h(x,y)}\big(\sigma^\perp_{1}(x,y,z)-\sigma^\perp_{2}(x,y,z)\big)\,dz,
\label{eq:Rt}
\end{equation}
where $C_{so}$ is the stress--optic coefficient, $h(x,y)$ is the material thickness along the optical path, and $\sigma^\perp_{1},\sigma^\perp_{2}$ are the maximum and minimum principal stresses in the $x$--$y$ plane.

To achieve a compact form suitable for size-constrained deployment, MVP-Tac uses a \emph{reflective} circular polariscope configuration as shown in Fig.\ref{fig:sensor}(d). The light source and camera are placed on the same side of the sensing layer and work as follows: a light source illuminates the stack; the light passes through a linear polarizer and a quarter-wave plate; it then propagates through the photoelastic material and reflects off a reflective film; finally, the reflected light traverses the same stack in reverse and is captured by the camera. Since the light passes through the photoelastic material twice, the effective optical retardation is doubled:
\begin{equation}
R^{ref}_t(x,y)=2R_t(x,y).
\label{eq:Rref}
\end{equation}

This reflective layout increases sensitivity while enabling a compact, light-shieldable module, and the resulting interference pattern as in Eq.~\ref{eq:Rt}, recorded by the camera serves as the tactile measurement signal.

\subsection{Design and Fabrication}
\begin{figure*}[!t]
    \centering
    \includegraphics[width=1\textwidth]{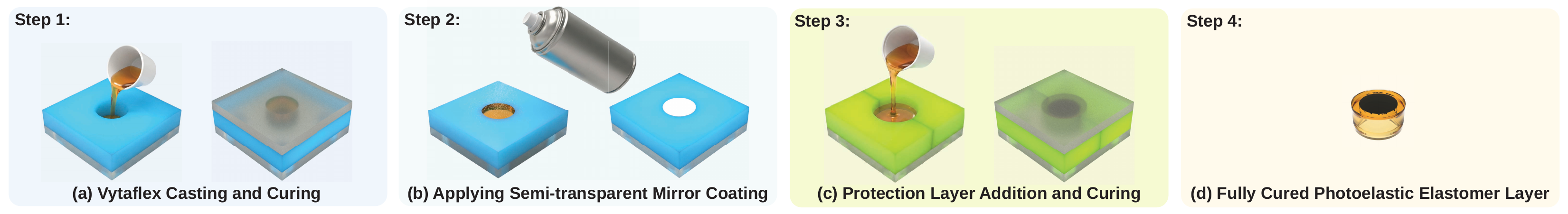}
    \caption{Fabrication process of the photoelastic elastomer layer}
    \label{fig:fab}
\end{figure*}

In this section, we describe the complete design and fabrication process of MVP-Tac. It employs a novel combination of photoelastic elastomer layers and a semi-transparent membrane to combine tactile sensing with vision modality in a compact and easily applicable form, eliminating the need for structured lighting and bulky arrangement. The sensor consists of three key elements: the dual-mode elastomer module (1), the polariscope optical module (2), and the perception module (3), as shown in Fig.\ref{fig:sensor}(c). The primary components and fabrication molds are easily accessible, off-the-shelf materials or are produced by 3D printing (Bambu Lab X1C FDM 3D printer), as depicted in Fig.\ref{fig:fab}. The following subsections provide a detailed explanation of each module and its fabrication procedure.

\subsubsection{Dual-mode Elastomer Module}

The elastomer module mainly comprises the fabrication of two separate but integrated layers: a photoelastic elastomer layer and a semi-transparent coating layer. The fabrication process of this module follows a structured procedure to ensure consistency and precision in tactile response. 

\paragraph{Photoelastic Elastomer Layer}
We use VytaFlex\textsuperscript{\texttrademark}~20 (Smooth-On, Inc.), an amber polyurethane rubber, as the photoelastic material. It is prepared by mixing Part~A and Part~B at a 1:1 ratio by weight, followed by a thorough mixing and degassing until the bubbles disappear. The mixture is then poured into a 3D-printed mold (5~mm diameter, 3~mm thickness) as shown in Fig.~\ref{fig:fab}(a) and pressed with an acrylic plate to achieve an optically smooth and flat surface and cured for 16~hours at room temperature. 

\paragraph{Semi-transparent Layer}
The semi-transparent membrane enables MVP-Tac to alternate between vision and tactile sensing by adjusting the interior--exterior brightness. With the LED arrays off (exterior brighter), the sensor operates in \emph{visual mode} and the camera sees through the membrane; with the LEDs on (interior brighter), it operates in \emph{tactile mode} and the camera records tactile imprints as shown in Fig.~\ref{fig:sensor}(e).

After curing the VytaFlex layer, we apply 2--3 coats of mirror-silver coating (Rust-Oleum Specialty Mirror Spray), as shown in Fig.~\ref{fig:fab}(b) to produce a semi-transparent (half-silvered) membrane, whereas applying 5--6 coats produces an opaque reflective layer. A thin VytaFlex overlayer is then cast in a relatively bigger mold (7~mm diameter, 3.5~mm thickness) as shown in Fig.~\ref{fig:fab}(c) and cured for 16~hours to protect the coating. Heat curing is avoided to prevent hardness changes, bubble formation, and surface tackiness. 

We cure the membrane against a smooth mold surface to avoid imprints and apply the spray coating uniformly, keeping the protective overlayer thin to maintain high-fidelity tactile imprints. Minimizing surface defects during fabrication helps preserve transparency and imprint quality. 
\subsubsection{Polariscope Optical Module}
As shown in Fig.~\ref{fig:sensor}(d), MVP-Tac employs a reflective circular-polariscope configuration. A linear polarizer first converts the LED illumination into linearly polarized light. A quarter-wave retarder then introduces a $\lambda/4$ phase delay between orthogonal polarization components (due to its different refractive indices along two principal axes), converting the light into circular polarization.

We use a left-handed circular polarizing film (CP42HE, Edmund Optics), which integrates the linear polarizer and quarter-wave retarder, mounting it directly above the camera and LED array as shown in Fig.~\ref{fig:sensor}(b). During assembly, the \emph{linear polarizer side} should be oriented to face the camera; otherwise, the photoelastic response is significantly reduced.

A 10~mm-thick acrylic spacer is added to satisfy the camera's working distance and maintain focus, while also providing mechanical support for the elastomer layer. The circular polarizer and acrylic spacer are fixed inside the housing using a UV-cured epoxy.

\subsubsection{Perception Module}

The perception module is the core imaging and illumination system of MVP-Tac. It uses an off-the-shelf miniaturized camera module (Guangzhou Sincere Info Tech) equipped with an endoscopic CMOS image sensor, offering a resolution of $1280 \times 720$ pixels, ideal for capturing detailed images necessary for accurate analysis. Considering the ultra-compact size requirements of our sensor, the camera module minimizes optical distortion while fitting within the limited inner cross-sectional area of $\pi \times 2.5^2$ $mm^2$ of the sensor, offering a 76° field of view (FOV) and a depth of field of $10 - 100~mm$. Fastened to the sensor base with M2 screws, it uses a standard USB-based camera, ensuring compatibility across computing platforms.

\subsubsection{Assembly}
The final assembly, shown in Fig.~\ref{fig:sensor}(b), integrates all three modules into a compact sensor. A key challenge was bonding the cured VytaFlex elastomer layer to the acrylic spacer, because VytaFlex (polyurethane) does not chemically bond to acrylic and the acrylic cross-section area is small. Directly casting VytaFlex onto the spacer produced a joint that could fail during calibration so to improve adhesion, we used Ure-Bond\textsuperscript{\texttrademark} II (Smooth-On, Inc.). The adhesive was prepared by mixing Part~A and Part~B in a 100:92 ratio by weight for 30~s, followed by degassing for 3~min. The cured VytaFlex layer as in Fig.~\ref{fig:fab}(d) was then bonded to the acrylic spacer. All bonding steps were completed within the $\sim$5~min pot life of Ure-Bond II, followed by a 5 to 6~hour cure for complete fabrication.

The entire process is simple and reproducible, with a compact yet adaptable design for RMIS and other applications requiring miniature tactile sensing. The total cost of sensor components and mold, including camera is under \$20.

\subsection{Sensor Calibration}
\begin{figure}[t]
    \centering
    \includegraphics[width=0.6\columnwidth, scale=0.7]{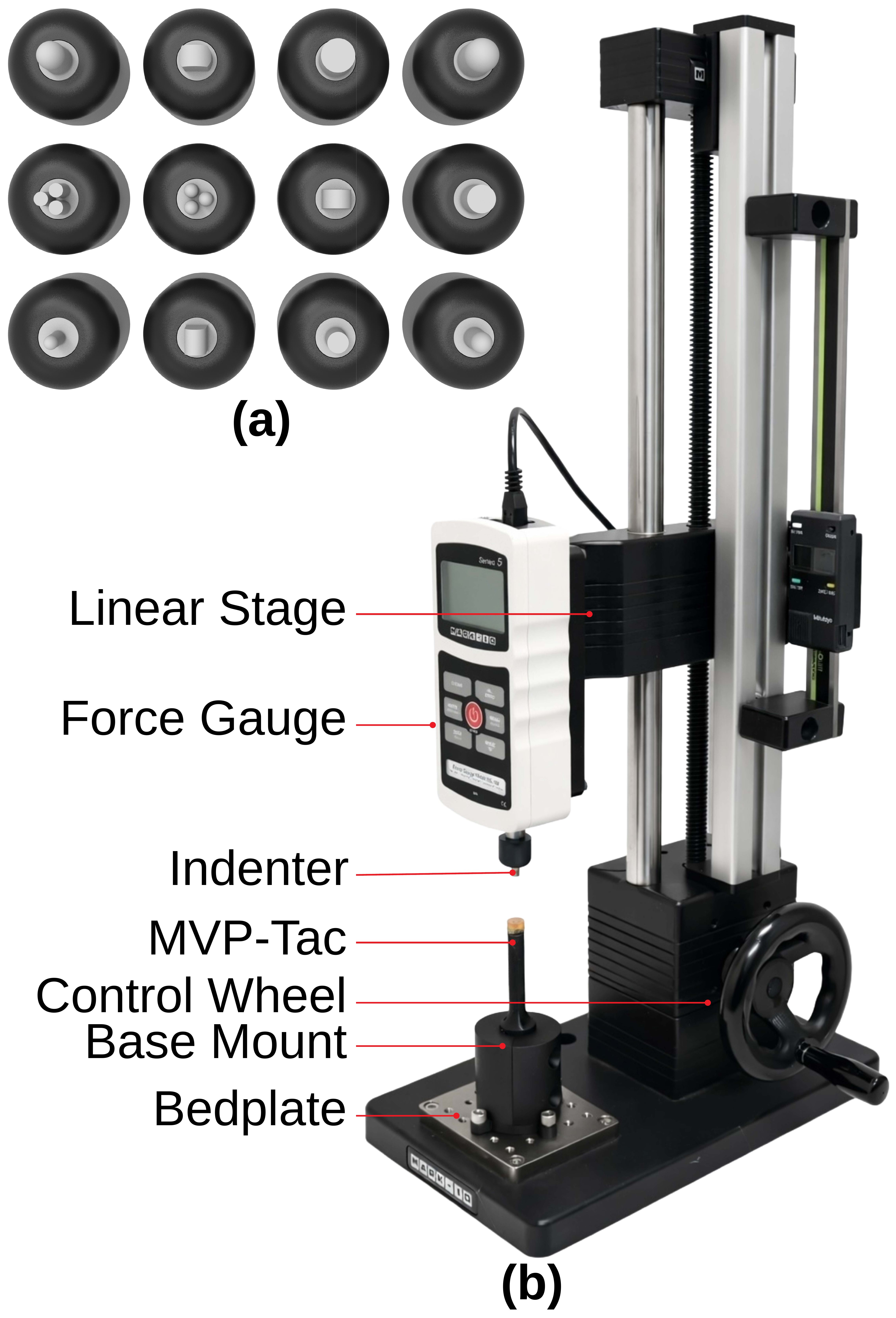}
    \caption{\textbf{Experimental setup for calibrating the MVP-Tac sensor.} 
A Mark-10 force gauge is mounted on a linear stage to apply controlled normal forces. The MVP-Tac sensor is fixed at the base of the stage, allowing precise force application and measurement during calibration.}
    \label{fig:calibration}
\end{figure}

MVP-Tac's force estimation is evaluated by measuring the applied normal force at contact. Motivated by tissue palpation (and other fine manipulation tasks), we focus on light forces up to 2~N. We probe the sensing surface using indenters of different shapes and sizes as shown in Fig.~\ref{fig:calibration}(a). As shown in Fig.~\ref{fig:calibration}(b), MVP-Tac is mounted on a base plate and an indenter tip is attached to a Mark-10 force gauge with a linear stage. During indentation, we synchronously record the contact images and normal-force readings, collecting approximately 2500 (image, force) pairs. Data were split into training/valid/test sets with a 70:10:20 ratio.

In addition to the raw images, we compute the photoelastic gradient metric $G^2$ from each captured frame as an auxiliary signal. Following the g-square method, the per-pixel value is defined as the squared intensity-gradient magnitude, and the overall $G^2$ for a region of interest (ROI) is obtained by averaging over all ROI pixels \cite{majmudar2006experimental}:
\begin{equation}
G^2=\frac{1}{N}\sum_{p\in\mathrm{ROI}}\left(\left(\frac{\partial I}{\partial x}\right)^2+\left(\frac{\partial I}{\partial y}\right)^2\right),
\label{eq:g2}
\end{equation}
where $I$ is the image intensity and $N$ is the number of pixels in the ROI. In our implementation, $I$ is taken from the green channel to compute $G^2$, as it provided a stable fringe contrast in our optical setup. We observe an approximately linear relationship between the average $G^2$ (averaged over the full sensing area) and the applied normal force as shown in Fig.~\ref{fig:response}(a). Therefore, both the computed $G^2$ feature and the raw tactile image are fused using a dual-branch neural network. The grayscale image is processed through a convolutional feature extractor to learn spatial deformation characteristics, while the scalar $G^2$ value is encoded through a lightweight fully connected branch to represent the global stress distribution. The resulting feature embeddings are concatenated and passed through fully connected layers to predict the applied force, allowing the network to exploit complementary local texture and global stress information.


Fig.~\ref{fig:response} shows that the fusion model accurately predicts forces across the 0--2~N range. The estimated forces closely follow the identity line (Fig.~\ref{fig:response}(b)), and the absolute errors are concentrated near zero with few outliers (Fig.~\ref{fig:response}(c)). The model achieves an MAE of 0.0511~N, MSE of 0.004690, RMSE of 0.0685~N, and $R^2$ of 0.9874. The average error of $\sim$0.05~N is well below the 0.1--0.2~N force error commonly considered acceptable for RMIS force sensing~\cite{othman2022tactile,chua2023modular}, demonstrating reliable image-to-force calibration for contact-rich palpation tasks.





\subsection{Sensor Characterization}

\begin{figure*}[!t]
    \centering
    \includegraphics[width=\textwidth]{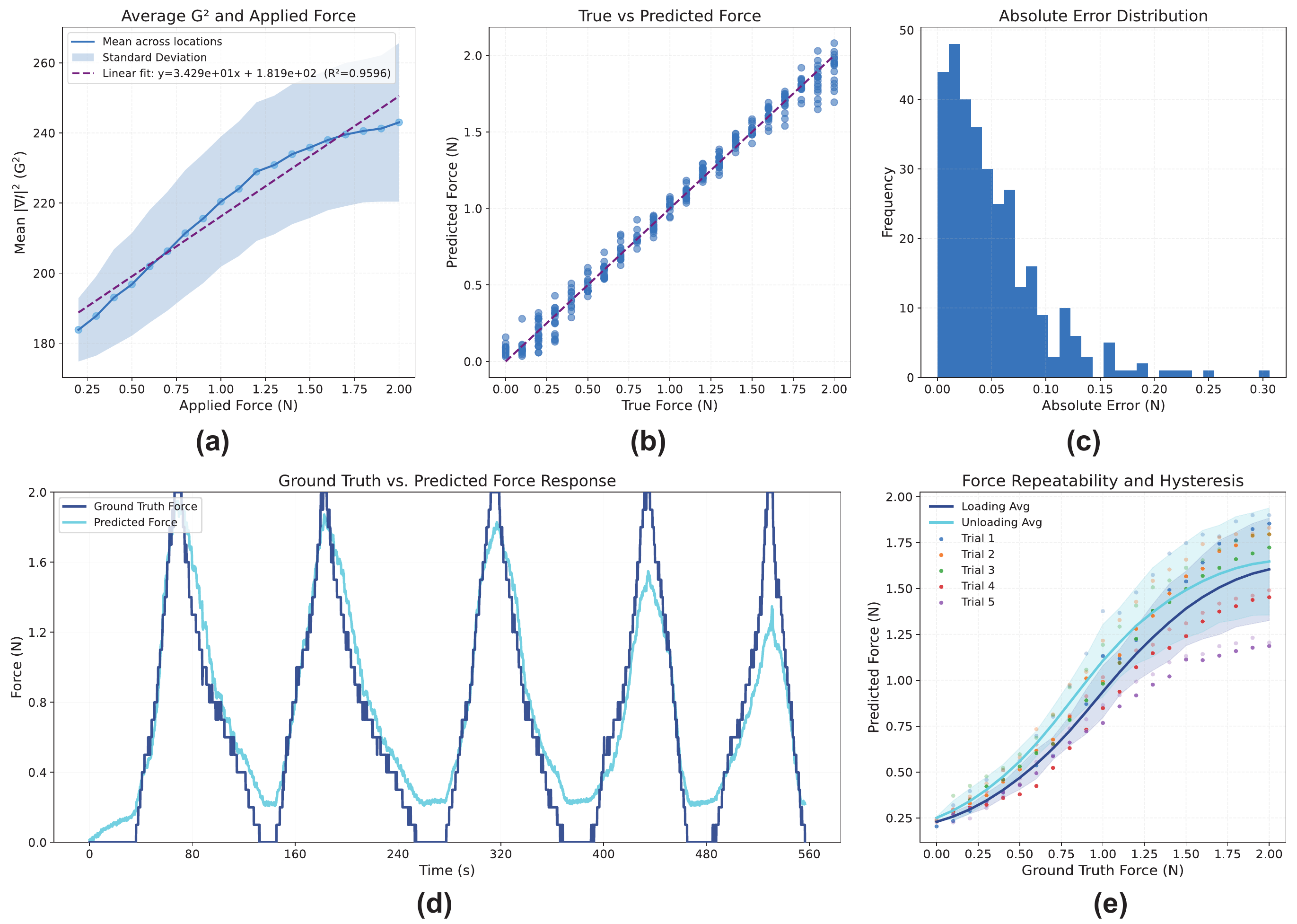}
    \caption{\textbf{Force repeatability and hysteresis characterization.}
Dots represent measurements from five repeated trials, while the solid curves indicate the smoothed trial-averaged responses during loading and unloading cycles.}
    \label{fig:response}
    \vspace{-10pt}
\end{figure*}

We characterize both sensing modalities of MVP-Tac. For the vision mode, we evaluate the spatial resolution of the optical imaging system and for the tactile mode, we characterize the repeatability and hysteresis of force estimation within the intended operating range (0 to 2\,N). The sensor is mounted on a Mark-10 force gauge, and five repeated loading--unloading trials are conducted up to the maximum applied ground-truth force of $F_M=2.0\,\mathrm{N}$. Figure~\ref{fig:response}(d) shows a representative time response of the ground-truth force and the model-predicted force over multiple cycles. To evaluate repeatability and hysteresis at matched force levels, we discretize the ground-truth force into 0.1\,N steps (with a $\pm$0.05\,N bin around each step) and compute statistics across trials.

\paragraph{Vision Spatial Resolution.}
The vision modality was characterized using a standard USAF 1951 resolution target as shown in Fig.~\ref{fig:colon2}(b). The smallest resolvable pattern was Group~1, Element~3, corresponding to 2.52 line pairs/mm, yielding an estimated spatial resolution of approximately 198~$\mu$m on the tactile sensing surface.

\paragraph{Repeatability}
Repeatability $r$ of MVP-Tac is computed from the maximum measurement spread at the same force step across all trials:
\begin{equation}
r = \frac{\Delta \hat{F}_{t_M}}{F_M}\times 100\%
  = 33.34\%.
\label{eq:repeatability_force}
\end{equation}
where $F_M=2.0\,\mathrm{N}$ is the maximum ground-truth force applied in this study, and $\Delta \hat{F}_{t_M}$ is the maximum difference between trials at a matched force step (i.e., the maximum spread of predicted force at the same ground-truth force bin).
In our implementation, we compute a per-trial mean predicted force within each force bin and then take the maximum spread across the five trials. 

\paragraph{Hysteresis}
Hysteresis quantifies the maximum discrepancy between the loading and unloading processes at the same force step. 
Using the smoothed trial-averaged loading and unloading curves in Fig.~\ref{fig:response}(d), we define
\begin{equation}
h = \frac{\Delta \hat{F}_{p_M}}{F_M}\times 100\%
  = 9.46\%.
\label{eq:hysteresis_force}
\end{equation}
where $\Delta \hat{F}_{p_M}$ is the maximum difference between the loading and unloading predicted responses at matched force steps. 
Since the loading and unloading curves do not perfectly overlap in Fig.~\ref{fig:response}(e), the observed hysteresis is attributed to the viscoelastic behavior of the elastomer and rate-dependent effects during cyclic loading.

Near the maximum load ($\sim$2N), the predictions begin to deviate from the ground truth for two expected reasons. First, the response becomes less linear as the elastomer approaches its compression limit, leading to geometric stiffening and partial saturation of the optical features, which reduces sensitivity at the upper end. Second, VytaFlex recovers slowly after deformation; under repeated cycles it may not fully return to its initial shape without sufficient dwell time, introducing residual strain and a phase lag that causes underprediction during rapid loading and overprediction during early unloading.

\section{EXPERIMENTS AND RESULTS}
To assess MVP-Tac’s ability to detect embedded abnormalities, we performed controlled experiments on tissue-mimicking phantoms. As tissue hardness is reflected in the photoelastic response under applied force, the demonstrated force sensing capability forms the basis for reliable hardness classification. We therefore trained a machine learning model to differentiate inclusions of varying hardness and further validated the sensor in a simulated colonoscopy setting.

\subsection{Artificial Tumor Detection}

Real tumors exhibit varying stiffness, but prior tactile sensor evaluations often rely on rigid inclusions, limiting generalizability. To address this, we evaluated MVP-Tac using soft-tissue phantoms with silicone inclusions spanning multiple hardness levels, testing both exposed and subdermal configurations. Similar to prior work \cite{di2025usingfiberopticbundles}, we cast silicone inclusions with controlled hardness values of Shore 20, 30, 40, 50, and 60A to represent clinically relevant tissue stiffnesses. We validated hardness with a Shore A durometer by averaging five measurements per batch. To represent clinically relevant lesion sizes, we used 3D-printed molds with diameters of 4, 8, and 12 mm, yielding 15 total samples (three per hardness level).

\begin{table}[h]
    \centering
    \begin{tabular}{c|c|>{\centering\arraybackslash}p{2.5cm}}
        \hline
        \textbf{Hardness (Shore A)} & \textbf{Material Name} & \textbf{Ratio by Weight (Part A : Part B)} \\
        \hline
        20 & Vytaflex 20 & 1:1 \\
        30 & Mold Max 30 & 100:10 \\
        40 & Smooth-Sil 940 & 100:10 \\
        50 & Smooth-Sil 950 & 100:10 \\
        60 & Mold Max 60 & 100:3 \\
        \hline
    \end{tabular}
    \caption{Mixing ratios for soft tumor nodules}
    \label{tab:nodule}
\end{table}

\begin{figure}[htbp]
    \centering
    \includegraphics[width=\columnwidth]{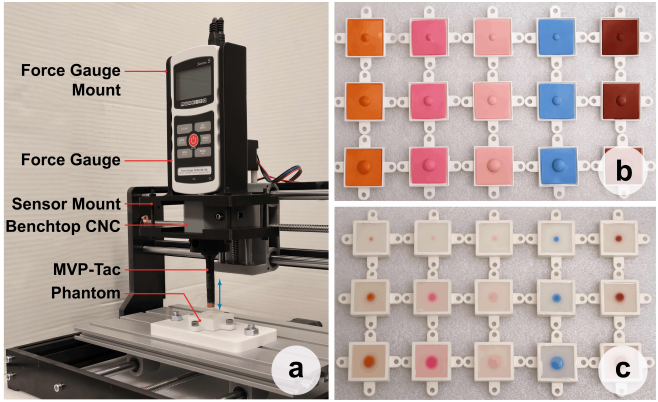}
    \caption{\textbf{Experimental setup for tumor detection}: (a) Tumor phantoms are placed on a CNC bed while MVP-Tac is mounted on the tool head to perform controlled indentation along the $z$-axis. (b) Exposed-tumor phantoms. (c) Subdermal-tumor phantoms. For both sets, hardness increases from 20A to 60A (left to right), and nodule diameter increases from 4 to 12,mm (top to bottom).}
    \label{fig:setup}
\end{figure}
\subsubsection{Exposed Tumors}
We first assessed MVP-Tac on exposed synthetic tumor nodules of varying hardness (Fig.~\ref{fig:setup}(b)). These phantoms were casted using 3D-printed molds, resulting in a rectangular base with the spherical phantoms atop both made of the same material (Table \ref{tab:nodule}). 


\subsubsection{Subdermal Tumors}

Next, we evaluated subdermal phantoms to represent plausible scenarios of non-exposed soft tumors. We inserted hemispherical tumor nodules atop 3 mm bases of Ecoflex 00-30, followed by another layer of Ecoflex 00-30 to fully submerge the tumors in 3D-printed molds, placing the nodules at a consistent depth of 2 mm beneath the surface as shown in Fig.~\ref{fig:setup}(c) to test MVP-Tac's ability to differentiate lesions of different hardness.

We mounted the samples on a CNC bedplate and attached MVP-Tac to the Mark-10 force gauge as shown in Fig.~\ref{fig:setup}(a). For each nodule, we performed repeated, gentle normal palpations (typically under 3 N), synchronously recording force and image sequences. The higher force limit compared to the 2 N force-calibration experiments was intentional, since soft subdermal inclusions required greater deformation of the overlying silicone layer to produce a measurable photoelastic response. To demonstrate MVP-Tac's palpation capabilities, we trained a video-based classifier to predict hardness from 16-frame sequences. Force measurements were recorded solely for reference and were excluded from classifier training.


We formulate tumor hardness recognition from palpation videos as a supervised sequence classification problem. Each palpation trial is recorded as an ordered RGB frame sequence; for every nodule, we collect 17 palpations at different spatial locations to capture variability in contact conditions. To prevent information leakage across temporally adjacent samples, we split the dataset at the trial level rather than at the frame level, yielding disjoint training, validation, and test sets.

Each trial is represented as a $T{=}16$-frame clip and we sample clips on random during training, using a deterministic center clip for validation/testing. Frames are resized to 160$\times$160, normalized, and augmented during training (random resized crop and mild color jitter), while evaluation uses center cropping only.

We use a lightweight ResNet-18+GRU architecture: ResNet-18 (ImageNet-pretrained) extracts a 512-D feature per frame, a GRU models temporal dynamics across the clip, and a linear head outputs the hardness class. The model is trained with cross-entropy loss using AdamW (learning rate $10^{-4}$), selecting the best checkpoint by validation accuracy. Training is performed on an NVIDIA A10 GPU (32~GB).

\begin{figure}[t]
    \centering
    \begin{subfigure}[t]{0.49\columnwidth}
        \centering
        \includegraphics[width=\linewidth]{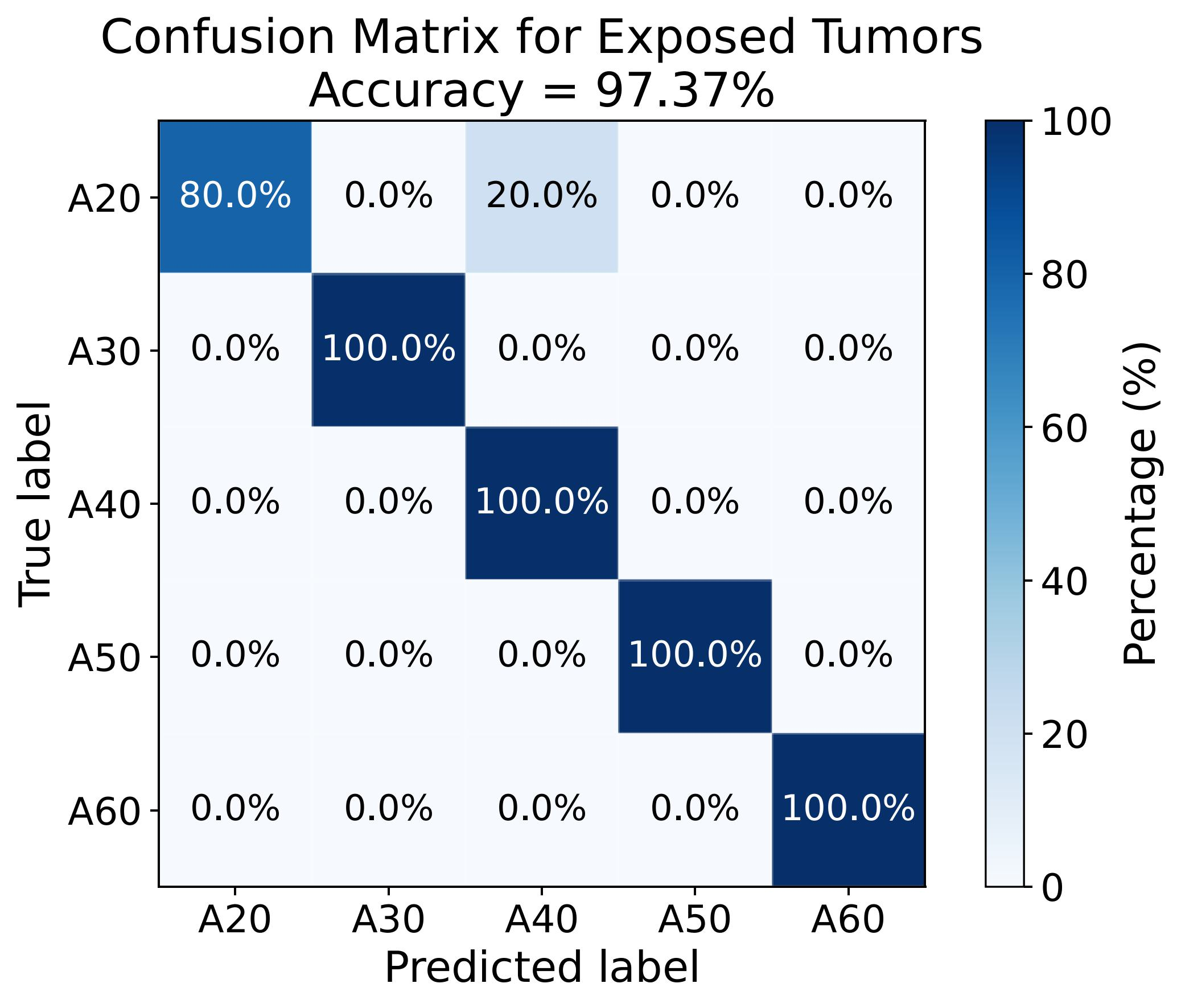}
        \caption{}
        \label{fig:cm_exposed}
    \end{subfigure}\hfill
    \begin{subfigure}[t]{0.49\columnwidth}
        \centering
        \includegraphics[width=\linewidth]{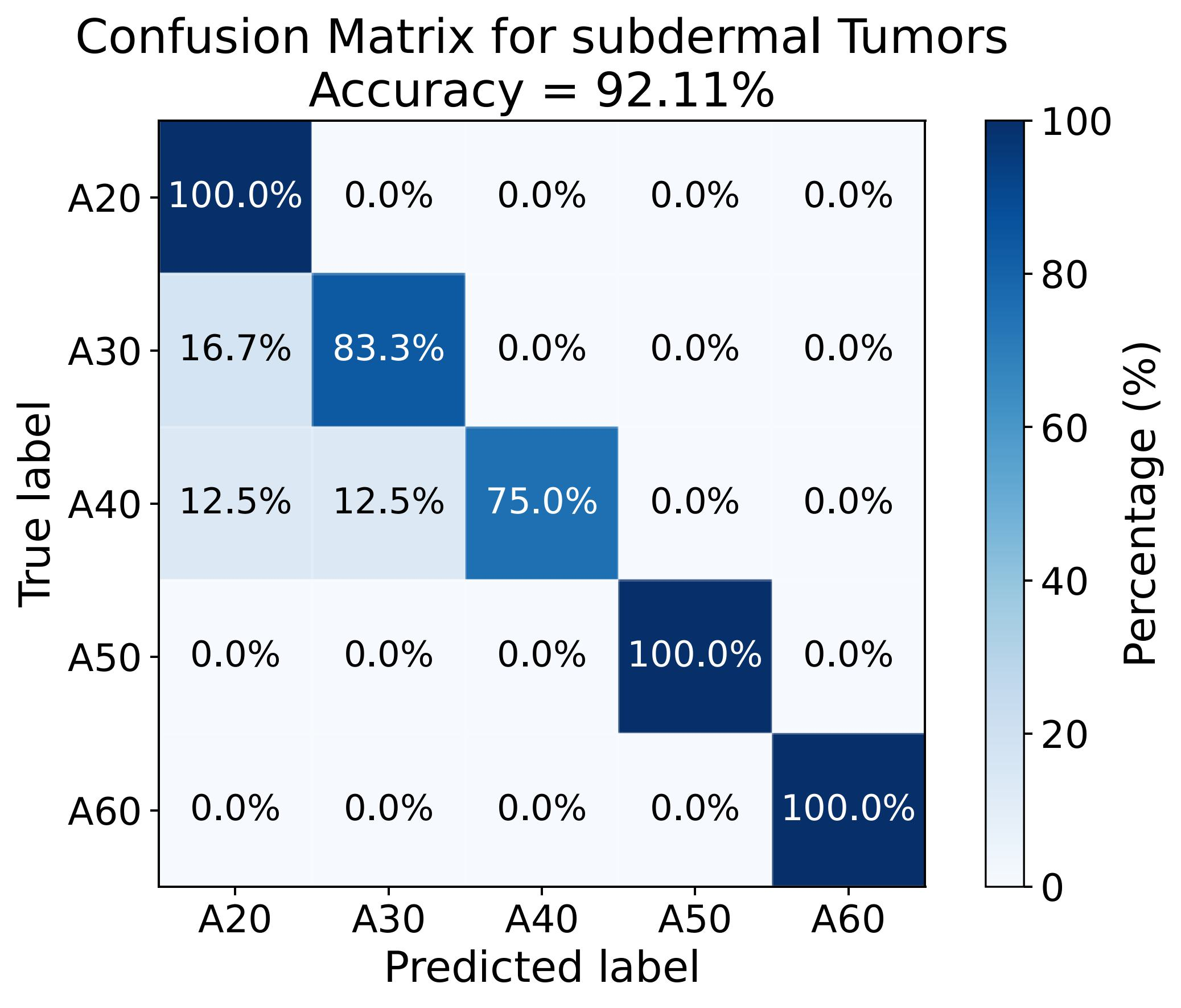}
        \caption{}
        \label{fig:cm_subdermal}
    \end{subfigure}

    \caption{\textbf{Confusion matrices for video-based hardness classification.}
    (a) Exposed tumor experiments and (b) Subdermal tumor experiments.}
    \label{fig:cm_both}
\end{figure}

On a held-out test set, the model achieves 97.37\% accuracy for exposed-tumor classification as in Fig.~\ref{fig:cm_both}(a) and 92.11\% for subdermal classification as in Fig.~\ref{fig:cm_both}(b). Real-time inference is demonstrated in the supplementary video.






\subsection{Simulated Colonoscopy with MVP-Tac}
\begin{figure}[htbp]
    \centering
    \includegraphics[width=0.8\columnwidth]{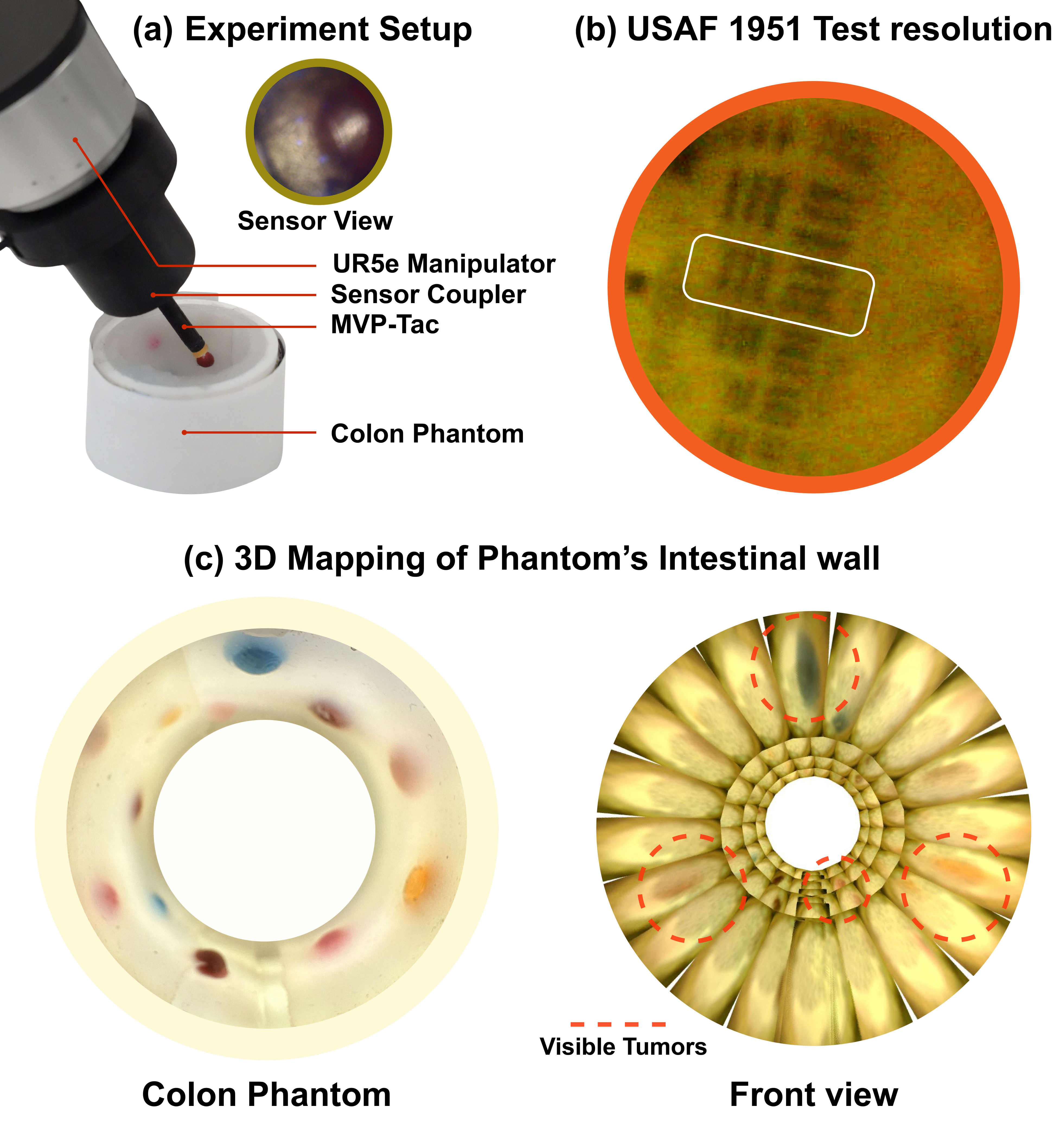} 
    \caption{\textbf{Colonoscopy setup experiment:} (a) Components of the experimental setup, with MVP-Tac gently palpating a tumor nodule. (b) Internal structure of the colon phantom. (c) Reconstructed 3D model of the phantom’s luminal wall generated from the sensor’s recorded visual stream.}
    \label{fig:colon2}
\end{figure}
\begin{figure}[htbp]
    \centering
    \includegraphics[width=\columnwidth]{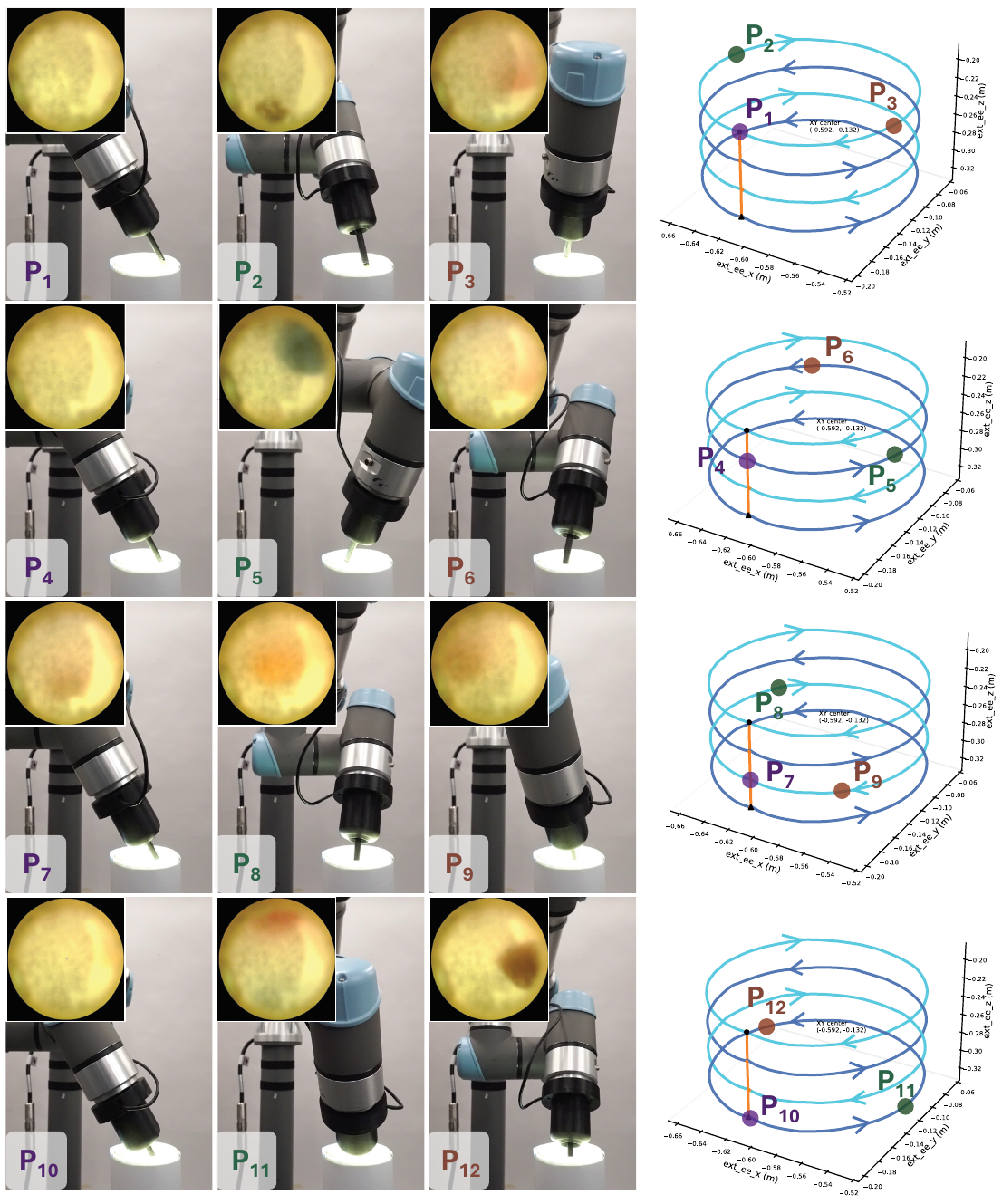}
    \caption{MVP-Tac mounted on a UR5e robot follows a stepwise circular scanning trajectory to map the luminal wall. The black region indicates the sensor’s camera view. After each full rotation, the manipulator advances 10~mm along the lumen and rotates in the opposite direction to achieve efficient coverage. During scanning, we record the video stream together with end-effector poses and robot joint angles, which are required for 3D mapping.}
    \label{fig:colon}
\end{figure}
To evaluate MVP-Tac in a clinically inspired scenario requiring continuous vision with intermittent palpation, we performed a simulated colonoscopy using an intestine-mimicking phantom consisting of a compliant cylindrical lumen (inner diameter $\sim$60~mm \cite{diameter}) with embedded artificial tumor nodules (Fig.~\ref{fig:colon2}(b)). The objectives were to evaluate visual navigation and nodule localization, followed by palpation-based confirmation of embedded abnormalities using the same sensor.

MVP-Tac was mounted on a UR5e end effector and inserted into the lumen, operating in \emph{visual mode} to survey the internal surface (Fig.~\ref{fig:colon}). The robot executed circular scans ($52\,\text{mm}$ diameter, $10\,\text{mm}$ layer spacing), while the robot states and sensor images were synchronously recorded at $10\,\text{Hz}$. The acquired images were downsampled to one-fifth resolution and fused into a 3D map of the lumen (Fig.~\ref{fig:colon2}(c)), enabling localization of candidate nodules based on their position, shape, and color. The sensor was then switched to \emph{tactile mode} to palpate 50A and 60A nodules under the lumen's space constraints, achieving 100\% hardness classification accuracy.

Overall, these results demonstrate MVP-Tac's ability to support a vision-guided \textit{search--then--palpate} workflow in a constrained lumen, consistent with RMIS and endoscopic screening.

\section{CONCLUSION AND FUTURE WORK}

This work introduced MVP-Tac, a compact vision-based tactile sensor that combines reflective photoelastic imaging with dual-mode operation for palpation in size-constrained RMIS settings. While its current sensing layer (VytaFlex) provides a strong photoelastic response, its Shore hardness ($\sim$20 A), early plastic deformation, slow recovery after loading, and fragility under highly concentrated contacts limit sensitivity, repeatability, and calibration. Future work will replace VytaFlex with biocompatible, softer photoelastic hydrogels to extend the detectable stiffness range and improve elastic recovery for reliable repeated palpation, while evaluating sterilization compatibility and robustness under physiological conditions and exposure to bodily fluids. MVP-Tac maintains an operating temperature of $\sim$24$^\circ$C, supporting safe contact with biological tissue. From an algorithmic standpoint, the current stress-to-force inference fuses a global scalar ($G^2$) with image features but does not fully exploit the spatial information in the interferogram. The next step would be to estimate dense force/stress fields directly from the image to enable richer tactile feedback beyond a single normal-force value. Finally, the current form factor is primarily limited by the camera module; ongoing miniaturization efforts target substantially smaller diameters for robust \textit{in vivo} deployment, potentially approaching the millimeter scale. 






\bibliographystyle{IEEEtran}

\bibliography{references}

@article{10.1145/2010324.1964941,
author = {Johnson, Micah K. and Cole, Forrester and Raj, Alvin and Adelson, Edward H.},
title = {Microgeometry capture using an elastomeric sensor},
year = {2011},
issue_date = {July 2011},
publisher = {Association for Computing Machinery},
address = {New York, NY, USA},
volume = {30},
number = {4},
issn = {0730-0301},
url = {https://doi.org/10.1145/2010324.1964941},
doi = {10.1145/2010324.1964941},
journal = {ACM Trans. Graph.},
month = jul,
articleno = {46},
numpages = {8}
}

@inproceedings{dong2017improved,
  title={Improved gelsight tactile sensor for measuring geometry and slip},
  author={Dong, Siyuan and Yuan, Wenzhen and Adelson, Edward H},
  booktitle={2017 IEEE/RSJ International Conference on Intelligent Robots and Systems (IROS)},
  pages={137--144},
  year={2017},
  organization={IEEE}
}

@article{othman2022tactile,
  title={Tactile sensing for minimally invasive surgery: conventional methods and potential emerging tactile technologies},
  author={Othman, Wael and Lai, Zhi-Han A and Abril, Carlos and Barajas-Gamboa, Juan S and Corcelles, Ricard and Kroh, Matthew and Qasaimeh, Mohammad A},
  journal={Frontiers in Robotics and AI},
  volume={8},
  pages={705662},
  year={2022},
  publisher={Frontiers Media SA}
}

@inproceedings{li2013sensing,
  title={Sensing and recognizing surface textures using a gelsight sensor},
  author={Li, Rui and Adelson, Edward H},
  booktitle={Proceedings of the IEEE Conference on Computer Vision and Pattern Recognition},
  pages={1241--1247},
  year={2013}
}

@inproceedings{luo2018vitac,
  title={Vitac: Feature sharing between vision and tactile sensing for cloth texture recognition},
  author={Luo, Shan and Yuan, Wenzhen and Adelson, Edward and Cohn, Anthony G and Fuentes, Raul},
  booktitle={2018 IEEE International Conference on Robotics and Automation (ICRA)},
  pages={2722--2727},
  year={2018},
  organization={IEEE}
}

@inproceedings{tian2019manipulation,
  title={Manipulation by feel: Touch-based control with deep predictive models},
  author={Tian, Stephen and Ebert, Frederik and Jayaraman, Dinesh and Mudigonda, Mayur and Finn, Chelsea and Calandra, Roberto and Levine, Sergey},
  booktitle={2019 International Conference on Robotics and Automation (ICRA)},
  pages={818--824},
  year={2019},
  organization={IEEE}
}

@inproceedings{wang2020swingbot,
  title={Swingbot: Learning physical features from in-hand tactile exploration for dynamic swing-up manipulation},
  author={Wang, Chen and Wang, Shaoxiong and Romero, Branden and Veiga, Filipe and Adelson, Edward},
  booktitle={2020 IEEE/RSJ International Conference on Intelligent Robots and Systems (IROS)},
  pages={5633--5640},
  year={2020},
  organization={IEEE}
}

@article{she2021cable,
  title={Cable manipulation with a tactile-reactive gripper},
  author={She, Yu and Wang, Shaoxiong and Dong, Siyuan and Sunil, Neha and Rodriguez, Alberto and Adelson, Edward},
  journal={The International Journal of Robotics Research},
  volume={40},
  number={12-14},
  pages={1385--1401},
  year={2021},
  publisher={SAGE Publications Sage UK: London, England}
}

@inproceedings{bauza2019tactile,
  title={Tactile mapping and localization from high-resolution tactile imprints},
  author={Bauza, Maria and Canal, Oleguer and Rodriguez, Alberto},
  booktitle={2019 International Conference on Robotics and Automation (ICRA)},
  pages={3811--3817},
  year={2019},
  organization={IEEE}
}

@inproceedings{suresh2022shapemap,
  title={Shapemap 3-d: Efficient shape mapping through dense touch and vision},
  author={Suresh, Sudharshan and Si, Zilin and Mangelson, Joshua G and Yuan, Wenzhen and Kaess, Michael},
  booktitle={2022 International Conference on Robotics and Automation (ICRA)},
  pages={7073--7080},
  year={2022},
  organization={IEEE}
}

@article{yuan2017gelsight,
  title={Gelsight: High-resolution robot tactile sensors for estimating geometry and force},
  author={Yuan, Wenzhen and Dong, Siyuan and Adelson, Edward H},
  journal={Sensors},
  volume={17},
  number={12},
  pages={2762},
  year={2017},
  publisher={MDPI}
}

@inproceedings{johnson2009retrographic,
  title={Retrographic sensing for the measurement of surface texture and shape},
  author={Johnson, Micah K and Adelson, Edward H},
  booktitle={2009 IEEE Conference on Computer Vision and Pattern Recognition},
  pages={1070--1077},
  year={2009},
  organization={IEEE}
}

@inproceedings{wang2021gelsight,
  title={Gelsight wedge: Measuring high-resolution 3d contact geometry with a compact robot finger},
  author={Wang, Shaoxiong and She, Yu and Romero, Branden and Adelson, Edward},
  booktitle={2021 IEEE International Conference on Robotics and Automation (ICRA)},
  pages={6468--6475},
  year={2021},
  organization={IEEE}
}

@inproceedings{taylor2022gelslim,
  title={Gelslim 3.0: High-resolution measurement of shape, force and slip in a compact tactile-sensing finger},
  author={Taylor, Ian H and Dong, Siyuan and Rodriguez, Alberto},
  booktitle={2022 International Conference on Robotics and Automation (ICRA)},
  pages={10781--10787},
  year={2022},
  organization={IEEE}
}

@inproceedings{romero2020soft,
  title={Soft, round, high resolution tactile fingertip sensors for dexterous robotic manipulation},
  author={Romero, Branden and Veiga, Filipe and Adelson, Edward},
  booktitle={2020 IEEE International Conference on Robotics and Automation (ICRA)},
  pages={4796--4802},
  year={2020},
  organization={IEEE}
}

@inproceedings{tippur2023gelsight360,
  title={Gelsight360: An omnidirectional camera-based tactile sensor for dexterous robotic manipulation},
  author={Tippur, Megha H and Adelson, Edward H},
  booktitle={2023 IEEE International Conference on Soft Robotics (RoboSoft)},
  pages={1--8},
  year={2023},
  organization={IEEE}
}

@article{xue20233,
  title={3-D dense reconstruction of vision-based tactile sensor with coded markers},
  author={Xue, Hongxiang and Sun, Fuchun and Yu, Haoqiang},
  journal={IEEE Transactions on Instrumentation and Measurement},
  volume={73},
  pages={1--8},
  year={2023},
  publisher={IEEE}
}

@article{du2021high,
  title={High-resolution 3-dimensional contact deformation tracking for fingervision sensor with dense random color pattern},
  author={Du, Yipai and Zhang, Guanlan and Zhang, Yazhan and Wang, Michael Yu},
  journal={IEEE Robotics and Automation Letters},
  volume={6},
  number={2},
  pages={2147--2154},
  year={2021},
  publisher={IEEE}
}

@inproceedings{yu2019novel,
  title={A novel tactile sensor based on structured light},
  author={Yu, Yuanlong and Xue, Hongxiang and Liang, Zhenzhen},
  booktitle={2019 IEEE International Conference on Robotics and Biomimetics (ROBIO)},
  pages={229--234},
  year={2019},
  organization={IEEE}
}

@article{li2024minitac,
  title={MiniTac: An Ultra-Compact 8 mm Vision-Based Tactile Sensor for Enhanced Palpation in Robot-Assisted Minimally Invasive Surgery},
  author={Li, Wanlin and Zhao, Zihang and Cui, Leiyao and Zhang, Weiyi and Liu, Hangxin and Li, Li-An and Zhu, Yixin},
  journal={IEEE Robotics and Automation Letters},
  year={2024},
  publisher={IEEE}
}

@inproceedings{lin2023dtact,
  title={Dtact: A vision-based tactile sensor that measures high-resolution 3d geometry directly from darkness},
  author={Lin, Changyi and Lin, Ziqi and Wang, Shaoxiong and Xu, Huazhe},
  booktitle={2023 IEEE International Conference on Robotics and Automation (ICRA)},
  pages={10359--10366},
  year={2023},
  organization={IEEE}
}

@article{lin20239dtact,
  title={9dtact: A compact vision-based tactile sensor for accurate 3d shape reconstruction and generalizable 6d force estimation},
  author={Lin, Changyi and Zhang, Han and Xu, Jikai and Wu, Lei and Xu, Huazhe},
  journal={IEEE Robotics and Automation Letters},
  volume={9},
  number={2},
  pages={923--930},
  year={2023},
  publisher={IEEE}
}

@article{athar2025vibtac,
  title={VibTac: A high-resolution high-bandwidth tactile sensing finger for multi-modal perception in robotic manipulation},
  author={Athar, Sheeraz and Zhang, Xinwei and Ueda, Jun and Zhao, Ye and She, Yu},
  journal={IEEE Transactions on Haptics},
  year={2025},
  publisher={IEEE}
}

@book{majmudar2006experimental,
  title={Experimental studies of two-dimensional granular systems using grain-scale contact force measurements},
  author={Majmudar, Trushant Suresh},
  year={2006}
}

@article{prince2025tacscope,
  title={TacScope: A Miniaturized Vision-Based Tactile Sensor for Surgical Applications},
  author={Prince, Md Rakibul Islam and Athar, Sheeraz and Zhou, Pokuang and She, Yu},
  journal={Advanced Robotics Research},
  pages={e202500117},
  year={2025},
  publisher={Wiley Online Library}
}

@book{aben2012photoelasticity,
  title={Photoelasticity of glass},
  author={Aben, Hillar and Guillemet, Claude},
  year={2012},
  publisher={Springer Science \& Business Media}
}

@article{ramesh2002digital,
  title={Digital photoelasticity: advanced techniques and applications},
  author={Ramesh, K and Lewis, G},
  journal={Appl. Mech. Rev.},
  volume={55},
  number={4},
  pages={B69--B71},
  year={2002}
}

@article{daniels2017photoelastic,
  title={Photoelastic force measurements in granular materials},
  author={Daniels, Karen E and Kollmer, Jonathan E and Puckett, James G},
  journal={Review of Scientific Instruments},
  volume={88},
  number={5},
  year={2017},
  publisher={AIP Publishing}
}

@article{puckett2013equilibrating,
  title={Equilibrating temperaturelike variables in jammed granular subsystems},
  author={Puckett, James G and Daniels, Karen E},
  journal={Physical Review Letters},
  volume={110},
  number={5},
  pages={058001},
  year={2013},
  publisher={APS}
}

@article{mitsuzuka2022application,
  title={Application of high-photoelasticity polyurethane to tactile sensor for robot hands},
  author={Mitsuzuka, Masahiko and Takarada, Jun and Kawahara, Ikuo and Morimoto, Ryota and Wang, Zhongkui and Kawamura, Sadao and Tajitsu, Yoshiro},
  journal={Polymers},
  volume={14},
  number={23},
  pages={5057},
  year={2022},
  publisher={MDPI}
}

@article{liu2025fatigue,
  title={Fatigue-Resistant Mechanoresponsive Color-Changing Hydrogels for Vision-Based Tactile Robots (Adv. Mater. 49/2025)},
  author={Liu, Jiabin and Li, Wei and She, Yu and Blanchard, Sean and Lin, Shaoting},
  journal={Advanced Materials},
  volume={37},
  number={49},
  pages={e71336},
  year={2025},
  publisher={Wiley Online Library}
}

@article{hu2025vibrissae,
  title={Vibrissae-inspired vision-based magnetic-actuated whisker},
  author={Hu, Zhixian and Cheng, Yi and Wachs, Juan and She, Yu},
  journal={Nature Communications},
  year={2025},
  publisher={Nature Publishing Group UK London}
}

@incollection{diameter,
  author    = {Ahmed, S. and Sharman, T.},
  title     = {Intestinal Pseudo-Obstruction},
  booktitle = {StatPearls [Internet]},
  publisher = {StatPearls Publishing},
  address   = {Treasure Island (FL)},
  year      = {2025},
  month     = {January},
  note      = {Updated 2023 Jul 3},
  url       = {https://www.ncbi.nlm.nih.gov/books/NBK560669/},
  urldate   = {2026-03-06}
}

@article{chua2023modular,
  title={A modular 3-degrees-of-freedom force sensor for robot-assisted minimally invasive surgery research},
  author={Chua, Zonghe and Okamura, Allison M},
  journal={Sensors},
  volume={23},
  number={11},
  pages={5230},
  year={2023},
  publisher={MDPI}
}

@misc{di2025usingfiberopticbundles,
      title={Using Fiber Optic Bundles to Miniaturize Vision-Based Tactile Sensors}, 
      author={Julia Di and Zdravko Dugonjic and Will Fu and Tingfan Wu and Romeo Mercado and Kevin Sawyer and Victoria Rose Most and Gregg Kammerer and Stefanie Speidel and Richard E. Fan and Geoffrey Sonn and Mark R. Cutkosky and Mike Lambeta and Roberto Calandra},
      year={2025},
      eprint={2403.05500},
      archivePrefix={arXiv},
      primaryClass={cs.RO},
      url={https://arxiv.org/abs/2403.05500}, 
}

\end{document}